\documentclass{article} 
\usepackage{preprint}
\usepackage{times}
\usepackage{multibib}

\usepackage[utf8]{inputenc} % allow utf-8 input
\usepackage[T1]{fontenc} % use 8-bit T1 fonts
\usepackage{hyperref} % hyperlinks
\usepackage{url} % simple URL typesetting
\usepackage{booktabs} % professional-quality tables
\usepackage{amsfonts} % blackboard math symbols
\usepackage{nicefrac} % compact symbols for 1/2, etc.
\usepackage{microtype} % microtypography
\usepackage{wrapfig}

\usepackage{rotating}
\usepackage{graphicx}
\usepackage{subfigure}
\usepackage{booktabs} % for professional tables
\usepackage{cleveref}
\usepackage{amsmath}
\usepackage{amssymb}
\usepackage{mathtools}

\usepackage{caption}
\title{IROF: a low resource evaluation metric for explanation methods}

\author{Laura Rieger \& Lars Kai Hansen \\
	DTU Compute\\
	 Technical University Denmark\\
	Lyngby, Denmark \\
	\texttt{\{lauri, lkai\}@dtu.dk} 
}

\iclrfinalcopy % Uncomment for camera-ready version, but NOT for submission.
\begin{document}
\maketitle

\begin{abstract}
	The adoption of machine learning in health care hinges on the transparency of the used algorithms, necessitating the need for explanation methods.
	However, despite a growing literature on explaining neural networks, no consensus has been reached on how to evaluate those explanation methods.
	We propose IROF, a new approach to evaluating explanation methods that circumvents the need for manual evaluation. Compared to other recent work, our approach requires several orders of magnitude less computational resources and no human input, making it accessible to lower resource groups and robust to human bias. 
\end{abstract}

\section{Introduction}

Health AI has already started to show value in high-income countries and there is broad consensus that the potential in low and middle income countries is even more significant \cite{wahl2018artificial}. Health AI will play important roles, e.g., for cost reduction, scaling of treatments and for training of new medical professionals. Explainability of Health AI is a critical factor for all of these dimensions. Explainability methods have matured considerably and numerous schemes have appeared for e.g. explaining image based diagnostics \cite{samekexplainable}, however, the evaluation of visual explanation methods remains an unsolved problem. Hence it remains a challenge to select among the many proposed explainability schemes and understand the value for a particular application. When designing an evaluation metric for explanation methods, certain desiderata apply, e.g. a metric should be quantitative in order to assist decision making, it should be sensitive to the important features of the give problem, and it should not be too costly to evaluate. 

We introduce IROF (\textbf{I}terative \textbf{R}emoval \textbf{O}f \textbf{F}eatures) as a new approach to quantitatively evaluate explanation methods without relying on human evaluation. The new evaluation metric is relevant to the features of medical decision problems as it is based on diagnostic accuracy and circumvents the problem of high correlation between neighbour pixels as well as the human bias that are present in current evaluation methods. The new method is computationally 'lightweight' compared to other proposed metrics.

\section{Previous work on evaluating explanation methods}
\label{sec:rel}

The evaluation of explanation methods is a relatively recent topic with few systematic approaches \citep{Ancona2017,hooker2018evaluating,Adebayo,Fong}. 
To our knowledge, \citet{Bach2015} proposed the first quantitative approach to evaluate an explanation method by flipping pixels to their opposite and comparing the decrease in output with the relevance attributed to the pixel by the explanation method. As the authors note, this only works for low-dimensional input.
 This work was followed up upon in \cite{samek2016evaluating}. By dividing high-dimensional images into squares, they make the method feasible for high-dimensional inputs. Squares with high relevance (as measured by the explanation method) consecutively get replaced with noise sampled from the uniform distribution. The difference between the original output and the output for the degraded images indicates the quality of the explanation method.

\citet{Ancona2017} proposed a different approach to evaluate explanation methods, called {Sensitivity-\textit{n}}, based on the notion that the decrease in output when several inputs are cancelled out should be equal to the sum of their relevances. As this is based on the assumption that input dimensions are not correlated, it is questionable for high resolution medical images.
%For a range of $ n $ (between 1 and the total number of inputs) they sample a hundred subsets of the input. For each $ n $, the Pearson Correlation Coefficient (PCC) between the decrease in output, when the subset of features is removed,
%and the sum of their relevances is reported. The result is a curve of the PCC dependent on the percentage of the input being removed. For a good explanation method, the PCC will decrease slowly. 

\citet{hooker2018evaluating} proposes a quantitative approach to evaluate explanation methods, ROAR. For each explanation method, they extract the relevance maps over the entire training set. They degrade the training set by setting different percentages of the pixels with the highest relevance to the mean and retrain the network. Each retrained network is evaluated on the test set. The accuracy on the test set decreases dependent on the percentage of pixels set to the mean.
Compared to our method, ROAR requires more computational resources in the order of magnitudes. Following the authors suggestions requires retraining ResNet50 25 times for each method. As a rough estimate, evaluating all methods we considered with ROAR would take 241 days using eight GPUs \footnote{Training ResNet50 with 8 Tesla P100 GPUs takes 29 hours according to \cite{goyal2017accurate}}, making it infeasible for most research groups. Due to this we will not consider ROAR further in this report.

\section{Evaluating explanation methods quantitatively with IROF }

\label{sec:gray}
\begin{figure*}[thbp]
	
	\begin{center}
		\includegraphics[width=.8\linewidth]{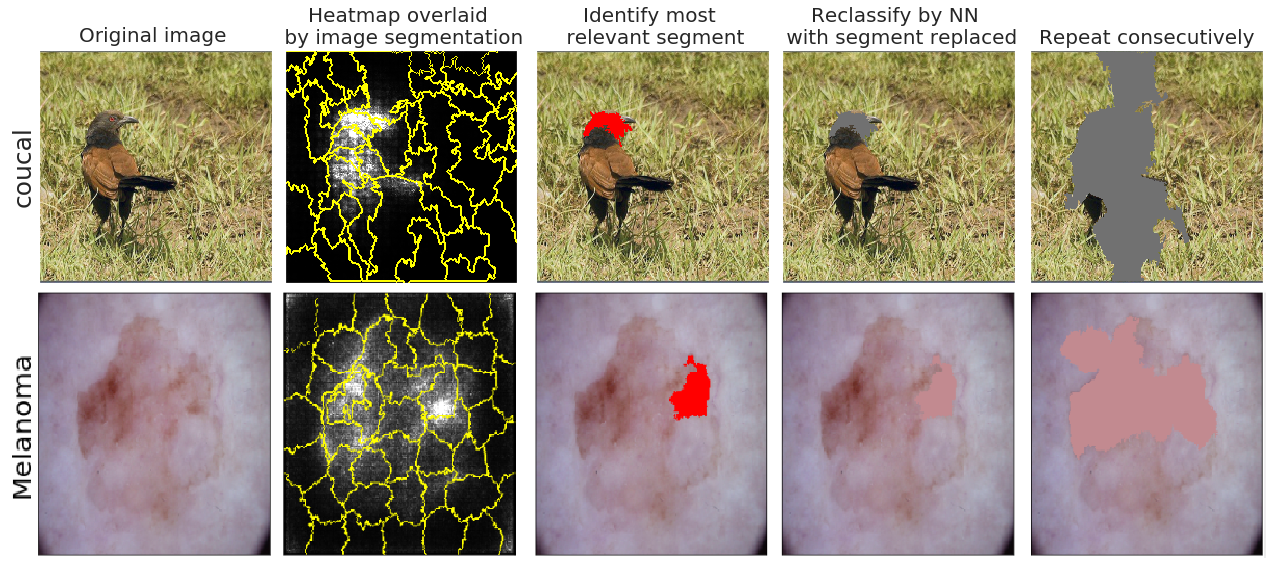}
	\end{center}
	\caption{Quantitative evaluation with IROF: Relevant segments as identified by an explanation method get consecutively replaced by the mean colour over the entire dataset. The IROF score of an explanation method is the integrated decrease in the class score over the number of removed segments. }
	\label{fig:graying_out}
	
\end{figure*}

Quantitative evaluation is a recurring problem with explainability methods. This is especially true for high-dimensional input, such as images, where important features consist of locally highly correlated pixels. If the information in one pixel is lost, this loss will not change the overall feature and should not result in a changed output score. The relevance values of single pixels are not indicative of the feature's importance as a whole.
We circumvent this problem by utilizing conventional image segmentation. First dividing the image into coherent segments bypasses the interdependency between the inputs. 

\paragraph{IROF methodology}

We assume a neural network $ F: X \mapsto y $ with $ X \in \mathcal{R}^{m \times m} $ and a set of explanation methods $ \{e_j\}_{j=1}^{J} $ with $ e_j : X,y,F \mapsto E$ with $ E \in \mathcal{R}^{m \times m} $. Without loss of generality we assume a quadratic image with width and height $ m $. 
Furthermore we partition each image $ X_{n} $ into a set of segments $ \{S_n^l\}_{l=1}^L $ using a given segmentation method with $ s_{n,i,j}^l = 1 $ indicating that pixel $ x_{n,i,j} $ belongs to segment $ l $. 
Computing the mean importance $ \frac{||E_{j,n} S_n^l||_1}{||S_n^l||_1} $ of each segment according to a given explanation method $ j $, two segments can be compared with each other. We then sort the segments in decreasing order of importance according to the explanation method.

 We use $ X_n^{'l} $ to indicate $ X_n $ with the $ l $ segments with highest mean relevance replaced with the mean value. Computing $ F(X_n^{'l})_y$ repeatedly with increasing $ l \in {0, ..., L} $ results in a curve of the class score dependent on how many segments of the image are removed. Dividing this curve by $ F(X_n^{'0})_y$ normalizes the scores to be within $ [0,1] $ and makes curves comparable between input samples and networks. 
A good explanation method will attribute high relevance to segments important for classification. As segments with high relevance are removed first, the score for the target class will decrease faster. By computing the area over the curve (AOC) for the class score curve and averaging over many input samples, we can score the methods according to how reliably they identify relevant areas of the image input. For a good explanation method, the AOC will be higher.
We refer to this evaluation method as the \textbf{iterative removal of features (IROF)}. The IROF score for a given explanation method $ e_j $ is expressed as:

\begin{equation}
	\text{IROF}(e_j) = \frac{1}{N} \sum_{n =1}^{N} \text{AOC} \left(\frac{F(X_n^{'l})_y}{F(X_n^{'0})_y} \right)_{l=0}^{L}
\end{equation}
This approach is a quantitative comparison of two or more explainability methods that does not rely on human evaluation or alignment between human and NN reasoning. For each explanation method the work-flow produces a single value, enabling convenient comparison between multiple explanation methods. A higher IROF score indicates that more information about the classification was captured.

IROF is dependent on having meaningful segments in the input, as natural images do. Dividing up text or non-natural images such as EEG into meaningful and independent segments does not have a natural solution and is left for future research.

\section{Experiments}
\label{sec:experiments}

We tested our method on five neural network architectures trained on ImageNet. Details are in the appendix. In addition we present results for a medical task, diagnosing skin diseases, in \cref{subsec:isic}.
We compared Saliency (SM), Guided Backpropagation (GB), SmoothGrad (SG), Grad-CAM (GC) and Integrated Gradients (IG) to have a selection of attribution-based methods \cite{Simonyan2013,Selvaraju,smilkov2017smoothgrad,Springenberg2014,Sundararajan2017a}. 
We use SLIC for image segmentation due to availability and quick run time \cite{achanta2012slic}.
Preliminary experiments with Quickshift showed similar results \citep{Vedaldi2008}. We therefore hypothesize that IROF is not sensitive to the choice of segmentation algorithm, although a more thorough study with more segmentation algorithms would be needed to confirm this. 

\subsection{Evaluating IROF for validity as an evaluation method}
\label{subsec:eval}
\begin{figure}[htb]
	
	\begin{center}
		\includegraphics[width=.9\linewidth]{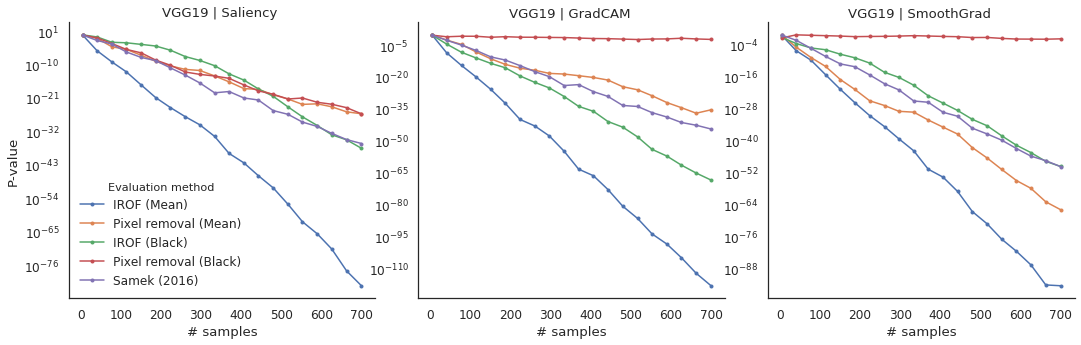}
	\end{center}
	\caption{P-values for the rejection of the random removal null-hypothesis. Lower is better. P-values are on a logarithmic scale. IROF performs best in all scenarios.}
	\label{fig:pstat}
	
\end{figure}

A good evaluation method should be able to reject the null hypothesis (a given explanation method is no better than random choice) with high confidence. Motivated by this, we compare IROF by calculating the paired t-test of an explanation method versus random guessing. This comparison is done with multiple explanation methods and networks, to reduce the impact of the explanation method.

We compare IROF and pixel removal with mean value and black as a replacement value respectively and additionally against \cite{samek2016evaluating}. 
For IROF and \cite{samek2016evaluating} we set the 10\% most relevant segments to the mean value over the dataset. For pixel removal, we set the equivalent number of pixels to the mean value. The percentage of segments or pixels being removed was chosen ad-hoc. 
If the difference in degradation between random choice and the explanation method is high, the explanation method reports meaningful information. Since we compare the same explanation method on the same neural network with different evaluation methods, the p-values only contain information about how meaningful the evaluation method is.

\begin{wraptable}{r}{0.5\textwidth}
	\caption{t-test: p-values of Random choice vs Saliency Mapping on forty images. All p-values $ < 0.05 $.}
	\label{tab:p_val}
	\begin{center}
		\begin{small}
			\begin{sc}
				\begin{tabular}{lrr}
					
					Evaluation method &  T-stat &     P-val \\
					\midrule
					IROF (Black) &    1.58 &  1.22e-01 \\
					IROF (Mean) &    5.44 &  3.60e-06 \\
					Pixel flipping (Black) &    1.92 &  6.22e-02 \\
					Pixel flipping (Mean) &    2.10 &  4.25e-02 \\
					Samek (2016) &    2.77 &  8.65e-03 \\
				\end{tabular}
			\end{sc}
		\end{small}
	\end{center}
\end{wraptable}
Results are shown in \cref{fig:pstat} (extended in \cref{sec:eval_eval}). In \cref{tab:p_val} we provide results for forty images in tabular form (other methods in \cref{sec:eval_eval}). On forty images, all evaluation methods produce p-values below 0.05. 
However, IROF can reject the null hypothesis (this explanation method does not contain any information), with much higher confidence with the same number of samples for any configuration. 
Thus, IROF is more sensitive to the explanation method than pixel removal or \cite{samek2016evaluating}, making it the better choice to quantitatively evaluate an explanation method.

Especially compared to \cite{samek2016evaluating}, IROF has several advantages. By taking natural image features into account via segmentation, it circumvents the problem of high local correlation between pixels belonging to the same feature. By normalizing the degraded outputs with the original output we decrease the dependency of IROF on the original outputs and consequently on the quality of the trained neural network and the clarity of the images. Furthermore, replacement with the mean value does not move an image as far from the input distribution as replacement with uniform noise, meaning that we do not measure how sensitive a neural network is to out-of-distribution input.
Thus, IROF is a more sensitive way to evaluate explanations for neural networks than pixel removal.
\subsection{Evaluating explanation methods with IROF}
\label{subsec:segment}

\begin{wrapfigure}{r}{0.5\textwidth}
	\begin{center}
		\includegraphics[width=\linewidth]{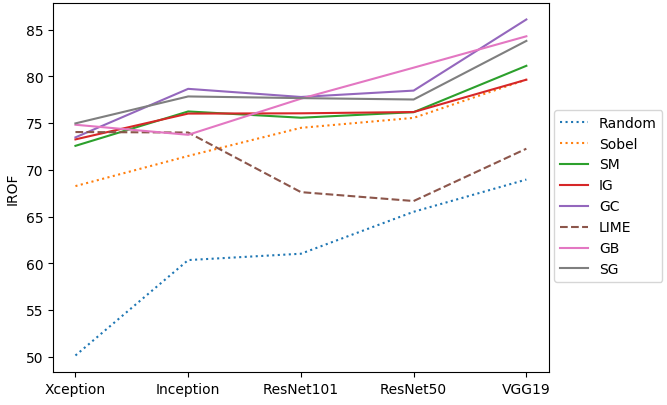}
	\end{center}
	\caption{IROF scores for different methods and networks. Higher is better. Dotted are baselines without any information about weights. The networks are ordered according to accuracy from high to low. }
	\label{fig:irof_scores_visualized}
\end{wrapfigure}
We apply IROF on multiple neural network architectures trained on ImageNet with a hundred randomly chosen correctly classified images from the test set and show the results in \cref{fig:irof_scores_visualized}. 
To check we are not simply measuring vulnerability of the neural network to degradation of the input we include two non-informative baselines, 
 \textit{Random} (randomly chooses segments to remove) and 
\textit{Sobel} (applying Sobel edge detection on the image and using the extracted edges as the relevance heatmap). 

 Additionally we compared against LIME as a method that is not based on attribution but on local approximation of the neural network \cite{Ribeiro}. 
The results in \cref{fig:irof_scores_visualized} (numbers in \cref{tab:results}) show several important insights:

\textbf{Explanation methods capture relevant information:}
All explanation methods have a higher IROF than the random baseline on all architectures tested. Except for LIME, all methods also surpass the stronger baseline, \textit{Sobel}, indicating that all backpropagation-based methods capture meaningful information on the classification.

\textbf{Explainability seems to be inversely correlated to accuracy:}
The IROF scores for all methods except LIME for a particular network architectures are strongly correlated. 
The networks in \cref{fig:irof_scores_visualized} are ordered according to accuracy from low to high. Higher IROF scores for a network, indicating easier interpretability, are correlated with lower accuracies. Especially in a medical context where transparency is important, this implies a trade-off between performance and interpretability. 
Measuring the interpretability of a neural network with IROF could provide an objective measure alongside accuracy.

\textbf{Explanation methods are relatively constant in their ordering:}
Though the individual IROF values vary, the ranking of explainability methods between networks is relatively constant, with GradCAM and SmoothGrad outperforming Integrated Gradients and Saliency Maps. We attribute the strong variance of LIME to the relatively high uncertainty due to random sampling that was already noted in \cite{zhang2019should}. 
As shown in \cite{Adebayo} the other exception, Guided Backprop (GP), is resistant to randomization of the neural network weights, supporting our finding here.
 In \cref{subsec:isic} we investigate whether this constant ordering is due to testing all methods on the same dataset or due to inherent properties of the explanation method.

\subsection{Evaluating explanation methods on ISIC skin cancer}
\label{subsec:isic}
\begin{table}[htb]
	\caption{IROF scores on a neural network trained for skin disease classification. Higher is better. }
	\label{tab:isic_irof}
	\begin{center}
		\begin{small}
			\begin{sc}
				
				\begin{tabular}{lrrrrrrrr}
					Method & Random 		& Sobel & LIME & SM& IG & GB& GC & SG\\
					\midrule
					Xception		& 19.2 & 		30.3 & 23.2 & 25.7 & 29.2 & 33.0 &35.2 &\textbf{35.3 }\\
				\end{tabular}
			\end{sc}
		\end{small}
	\end{center}
\end{table}
To investigate the validity of IROF  in a medical context, we consider the classification task of the ISIC skin lesion challenge \cite{ codella2019skin}. An Xception architecture is fine-tuned to classify seven skin diseases based on dermoscopic images (details in \cref{subsec:isic_detailed}). Results are shown in \cref{tab:isic_irof}. 
Despite the task being very different from ImageNet, the ranking of the respective methods is identical to the ImageNet task. SmoothGrad and GradCAM perform best, indicating that they would be most useful for aiding medical professionals. In contrast to the previous experiment, the Sobel edge detector outperforms both Saliency Mapping and Integrated Gradients. Given that Saliency Mapping is often used to check the validity of neural networks for medical tasks, this is an interesting result \cite{esteva2017dermatologist}. Especially in a low-resource environment, more reliable explanation methods can be used to aid cooperation between doctor and machine.

However, since we only finetuned the network weights with a small dataset, more experiments are needed to verify this result.
Additionally, IROF scores are substantially lower on the ISIC task than on the Imagenet task. This is not surprising, as it implies that diagnosis of skin lesions is harder to interpret than the classification of natural images.

\section{Conclusion}

The adoptions of health AI and realizing the huge potential in low and middle income countries is contingent on performance, usability and trust. In this context explainability is of paramount importance. However, the proposed schemes for explaining, e.g., visual diagnostics based on deep learning, currently come without a satisfying evaluation metric.

In this work we propose a novel way of evaluation for explanation methods that circumvents the problem of high correlation between pixels and does not rely on visual inspection by humans. To our knowledge, this is the first work that systematically evaluates the evaluation metric at hand. We found that IROF is more reliable than previous metrics while needing less resources. 
We evaluate several explanation methods on multiple architectures with IROF. The results imply a trade-off between accuracy and interpretability. We show evidence supporting the use of SmoothGrad (an aggregated version of Saliency Maps) over Saliency Maps. This was particularly  pronounced for medical diagnosis. 

Finally, we suggest the use of IROF for evaluating the inherent interpretability of a network architecture, giving practitioners an objective and intuitive metric to weigh against accuracy when deciding on neural network architectures for high-stakes tasks.

\clearpage
\bibliographystyle{irof}
\bibliography{irof}
\clearpage
\section{Appendix}

\subsection{Experimental setup }
\subsubsection{General}
We use SLIC for image segmentation due to availability and quick run time \cite{achanta2012slic}.
 Preliminary experiments with Quickshift showed similar results \citep{Vedaldi2008}. SLIC was chosen over Quickshift due to the quicker run time. The number of segments was set to $ 300 $ ad hoc. \cref{fig:graying_out} shows the same procedure with $ 100 $ segments for the sake of clarity. 

Some of the methods result in positive and negative evidence. We only conssdered positive evidence for the ImageNet tasks to compare methods against each other. To check that this does not corrupt the methods, we compared the methods that do contain negative results against their filtered version and found negligible difference between the two versions of a method in the used metrics.
\subsubsection{ImageNet}
We tested our method on five network architectures that were pre-trained on ImageNet: VGG19, Xception, Inception, ResNet50 and ResNet101 \cite{DengJ.andDongW.andSocherR.andLiL.-J.andLiK.andFei-Fei2009,simonyan2014very,he2016deep,chollet2017xception,szegedy2016rethinking}\footnote{Models retrieved from \href{https://github.com/keras-team/keras}{{https://github.com/keras-team/keras}}.}. Due to the softmax non-linearity commonly used in the last layer of neural networks, the output for all classes sum up to one, i.e.\ there is always at least one class with output greater than zero.

We downloaded the data from the ImageNet Large Scale Visual Recognition Challenge website and used the validation set only. No images were excluded. The images were preprocessed to be within $ [-1,1] $ unless a custom range was used for training (indicated by the preprocess function of keras).

The dataset consists of images from 1000 non-overlapping categories such as 'slug', 'pelican' or 'soccer ball'. Each image contains one and only one class. 

%On the set of images from the validation set of Imagenet we r

\clearpage
\subsection{Evaluating the evaluation}
\label{sec:eval_eval}
We report p-values for evaluating with 50 images on ResNet101 in the manner described in \cref{subsec:eval} in tabular form to provide a clear overview.
\begin{table}[!h]
	\caption{t-test p-values of explanation methods for SmoothGrad.}
	\label{tab:p_val_add}
	\vskip 0.15in
	\begin{center}
		\begin{small}
			\begin{sc}
				\begin{tabular}{lrr}
					
					Evaluation method &  T statistic &   P-value \\
					\midrule
					IROF (Black) &    3.97 &  3.22e-04 \\
					IROF (Mean) &    5.99 &  6.42e-07 \\
					Pixel flipping (Black) &    0.55 &  5.86e-01 \\
					Pixel flipping (Mean) &    5.00 &  1.39e-05 \\
					Samek (2016) &    2.97 &  5.17e-03 \\
				\end{tabular}
				
			\end{sc}
		\end{small}
	\end{center}
	\vskip -0.15in
\end{table}

\begin{table}[!h]
	\caption{t-test p-values of explanation methods for Saliency.}
	
	\vskip 0.15in
	\begin{center}
		\begin{small}
			\begin{sc}
				\begin{tabular}{lrr}
					\toprule
					Evaluation method &  T statistic &     P-value \\
					\midrule
					IROF (Black) &    3.97 &  3.22E-04\\
					IROF (Mean) &    5.99 &  6.42E-07 \\
					Pixel flipping (Black) &    0.55&  5.86E-01 \\
					Pixel flipping (Mean) &    5.00 &  1.39E-05 \\
					Samek (2015) &    2.97 &  5.17E-03 \\
					\bottomrule
				\end{tabular}
			\end{sc}
		\end{small}
	\end{center}
	\vskip -0.15in
\end{table}

\begin{table}[!h]
	\caption{t-test p-values of explanation methods for GradCAM.}
	
	\vskip 0.15in
	\begin{center}
		\begin{small}
			\begin{sc}
\begin{tabular}{lrr}
	
	Evaluation method &  T statistic &   P-value \\
	\midrule
	IROF (Black) &    4.73 &  3.23e-05 \\
	IROF (Mean) &    7.81 &  2.42e-09 \\
	Pixel flipping (Black) &   -1.75 &  8.85e-02 \\
	Pixel flipping (Mean) &    3.26 &  2.38e-03 \\
	Samek (2016) &    3.32 &  2.04e-03 \\
\end{tabular}

			\end{sc}
		\end{small}
	\end{center}
	\vskip -0.15in
\end{table}
\clearpage

We provide an extended version of \cref{fig:pstat}.
\begin{figure}[!h]
	\begin{center}
		\includegraphics[width=\linewidth]{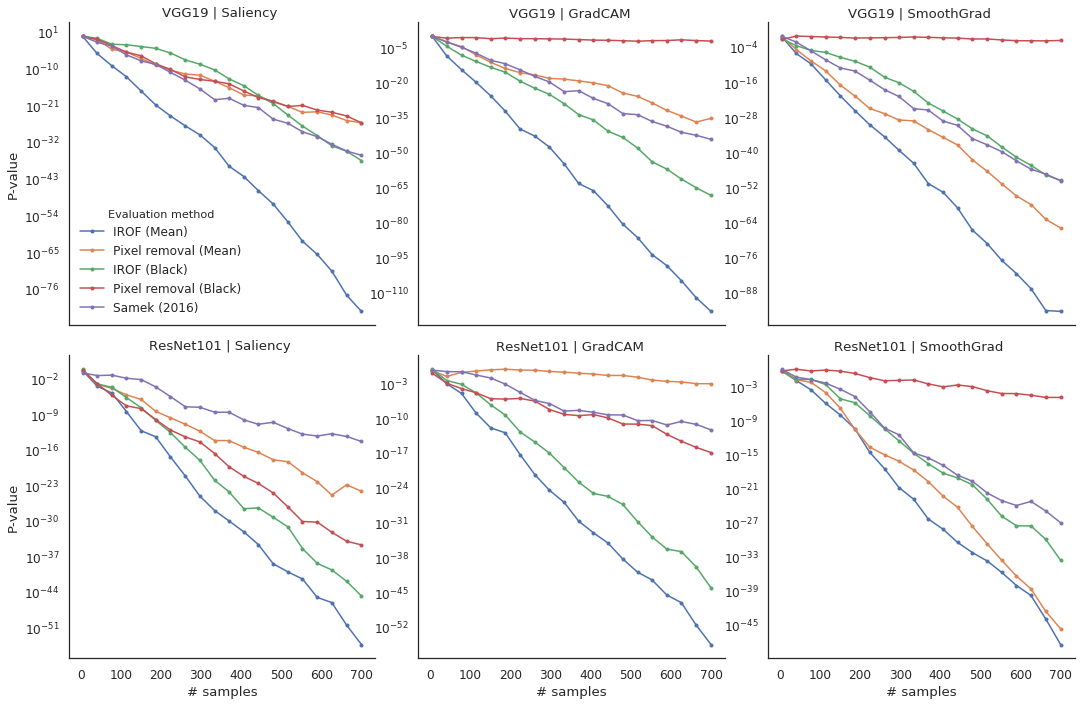}
	\end{center}
	\caption{P-values for the rejection of the random removal null-hypothesis. Extended graphs }
	\label{fig:pstat_supp}
	\vskip -0.2in
\end{figure}
\clearpage
\subsection{IROF scores for evaluation methods}
In the main script we visualize the IROF scores in a plot, since the comparison of eight methods over five networks is hard to comprehend in a table. We supply the numerical values in \cref{tab:results}.
\begin{table*}[ht]
	\caption{IROF scores across methods and architectures. All SE $< 0.05 $. }
	\label{tab:results}
	
	\begin{center}
		\begin{small}
			\begin{sc}\begin{tabular}{lrrrrr}
					Method              &     				Inception 		&  ResNet101 &  ResNet50 &  VGG19 &  Xception \\
					\midrule
					Random       			&       				60.3 &       		61.0 &      65.5 &   68.4 &      50.1 \\
					Sobel           	&       				71.5 &       		74.5 &      75.6 &   79.6 &      68.3 \\
					LIME             	&       				74.0 &       		67.6 &      66.7 &   73.4 &      74.0 \\
					SM                       &       			76.2 &       		75.6 &      76.2 &   81.4 &      72.6 \\
					%	GB                       &       			73.7 &       		77.6 &\textbf{80.9} &   84.3 &      74.8 \\
					IG                       &       			76.0 &       		76.0 &      76.2 &   79.8 &      73.3 \\
					SG                       &       			77.9 &       		77.7 &      77.5 &   83.9 &    \textbf{  75.0} \\
					GC                       &       		\textbf{78.7} & \textbf{77.8 }&      \textbf{78.5} &   \textbf{86.1} &      73.5 \\   
				\end{tabular}
				
			\end{sc}
		\end{small}
	\end{center}
\end{table*}
\subsection{ISIC 2018: Skin Lesion Analysis Towards Melanoma Detection }
\label{subsec:isic_detailed}
\subsubsection{Experimental details}
We trained an Xception architecture, using the pretrained weights and retraining the last layer. Since our goal was not to reach state of the art, we did not use any fine-tuning. On the test set, we reached 66.4\% accuracy (Macro ROC AUC: 0.81 ). 
The network is trained to distinguish seven disease categories: Melanoma,
Melanocytic nevus,
Basal cell carcinoma, 
Actinic keratosis / Bowen’s disease (intraepithelial carcinoma),
Benign keratosis (solar lentigo / seborrheic keratosis / lichen planus-like keratosis),
Dermatofibromam,
and Vascular lesion.
\subsubsection{Example images}
We show example images from the test set of \cite{ codella2019skin,tschandl2018ham10000}
\begin{figure}[ht]fs
	\centering
	
	\includegraphics[width=\linewidth]{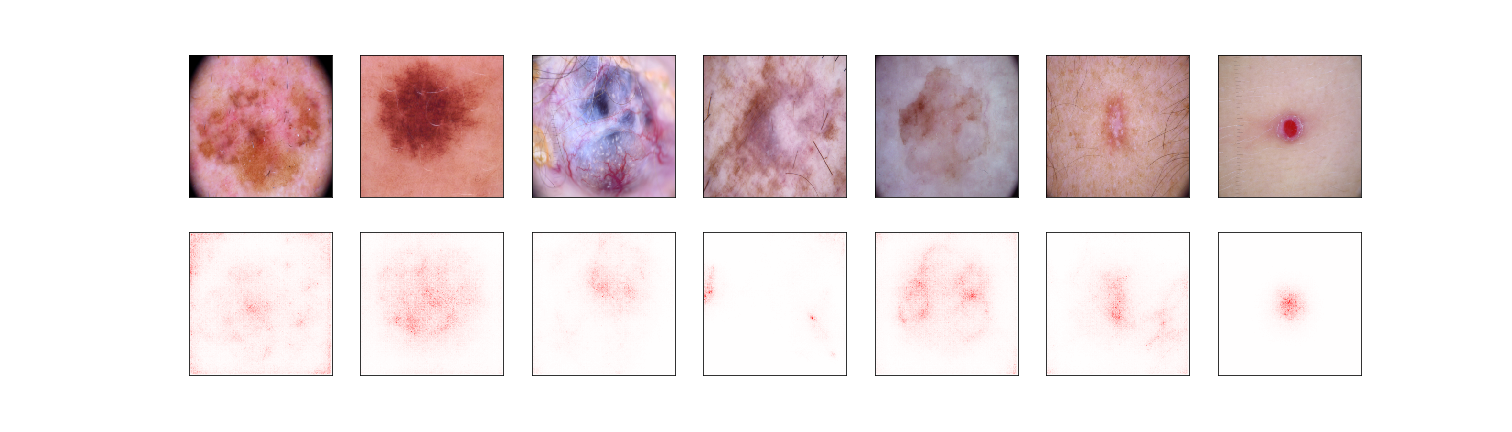}
	\caption{Upper row: Example images for the skin diseases. Lower row: SmoothGrad visualizations (best performing) for respective images. Lesions are marked as relevant. 
		From left to right: Melanoma,
		Melanocytic nevus,
		Basal cell carcinoma, 
		Actinic keratosis / Bowen’s disease (intraepithelial carcinoma),
		Benign keratosis (solar lentigo / seborrheic keratosis / lichen planus-like keratosis),
		Dermatofibromam,
		and Vascular lesion. }

	\label{fig:isic_example_imgs}
	
\end{figure}

\end{document}